\documentclass{article} % For LaTeX2e
\usepackage[final]{colm2026_conference}

\usepackage{microtype}
\usepackage{hyperref}
\usepackage{url}
\usepackage{booktabs}

\usepackage{comment}
\usepackage{graphicx}
\usepackage{makecell}
\usepackage{tabularx}
\usepackage{array}
\usepackage{amsmath}
\usepackage{float}
\usepackage{multirow}
\usepackage{bm}
\usepackage{subcaption}
\usepackage{wrapfig}
\usepackage[table]{xcolor}
\usepackage{colortbl}

\definecolor{blue1}{RGB}{219, 234, 254}
\definecolor{blue2}{RGB}{147, 197, 253}
\definecolor{blue3}{RGB}{59, 130, 246}
\definecolor{blue4}{RGB}{29, 78, 216}
\definecolor{blue5}{RGB}{30, 27, 150}

\newcommand{\best}[1]{\cellcolor{gray!20}\textbf{#1}}

\usepackage{lineno}

\definecolor{darkblue}{rgb}{0, 0, 0.5}
\hypersetup{
    colorlinks=true,
    citecolor=darkblue,
    linkcolor=darkblue,
    urlcolor=darkblue
}

\title{Knowing Before Answering: Decoding Language Models for Reliable RAG}

\author{
Syed Mahbubul Huq\textsuperscript{1},
Chris Child\textsuperscript{1},
Tillman Weyde\textsuperscript{1} \&
Pranava Madhyastha\textsuperscript{1,2} \\
\textsuperscript{1} City St George's, University of London \\
\textsuperscript{2} The Alan Turing Institute \\
\textbf{Correspondence: }\texttt{\{syed-mahbubul.huq.2,pranava.madhyastha\}@city.ac.uk}
}

\begin{document}

\ifcolmsubmission
\linenumbers
\fi

\maketitle

\begingroup
\renewcommand{\thefootnote}{}
\makeatletter
\renewcommand{\@makefntext}[1]{\noindent\raggedright #1}
\makeatother
\footnotetext{Code and dataset are available at
\url{https://github.com/SyedHuq28/Knowing-Before-Answering-Decoding-Language-Models-for-Reliable-RAG}.}
\endgroup

\begin{abstract}
In Retrieval-Augmented Generation (RAG), retrieval may provide insufficient or conflicting information needed to answer a question. The system should not only know when to answer but also be able to identify cases in which the documents provided in RAG are insufficient or contain conflicting information. This can be framed as a three-way classification problem, where we use the model’s internal signals to determine whether the provided information in the input can be classified as sufficient, insufficient, or conflicting. We create a controlled benchmark dataset that replicates a RAG setup with fictitious information and labels each instance as answerable, insufficient, or conflicting. We use hidden activations and attention-derived features as inputs to train a lightweight linear model to distinguish among the three classes. Across 16 language models spanning different architectures and a range of model sizes, our feature-based router consistently outperforms prompting-based baselines and the performance of specialised RAG-models. We further conduct analyses into the information dynamics of the models. We show that the most informative signals for the classification are available in the middle layers, with hidden activation states being more effective than attention values or the MLP-feature outputs in most of the tested models. Overall, our results suggest that language models internally encode whether retrieved evidence is sufficient to support answering, and that this signal can be decoded reliably for RAG triage.
\end{abstract}

\section{Introduction}

Retrieval-Augmented Generation (RAG) extends 
language models with external evidence at 
inference time, improving factuality and 
reducing hallucination \citep{Lewis2020RAG}. 
But retrieval is imperfect. In practice, 
retrieved documents may lack the information 
needed to answer a question, or may contain 
mutually contradictory answers. Prior work 
shows that in these cases, models tend to 
generate unsupported or contradictory 
responses rather than acknowledging the 
limitations of the evidence 
\citep{Niu2024RAGTruth, Joren2025SufficientContext}.

Existing methods address this as a two-way 
problem that involves either answering the question or refusing to answer the question. These predominantly rely on prompting \citep{Zhang2025FaithfulRAG, Zhou2023ContextFaithful}, confidence-based abstention \citep{Kadavath2022KnowWhatTheyKnow}, retrieval-side relevance estimation \citep{Yoran2024RetRobust}, or external entailment \citep{Honovich2022TRUE}. These approaches share two limitations: they operate on the output side of the model, and they collapse the problem to two classes, ignoring the distinct case where conflicting evidence is present. A model that confidently answers from contradictory documents is failing in a qualitatively different way from one that answers without any supporting evidence. We highlight here that this distinction is an important one, especially for downstream tasks, trust and system design. 

In this work, we propose a three-way framing: for a 
given question and retrieved context, the 
evidence state is either \emph{Answer} 
(sufficient, consistent evidence), 
\emph{Refuse} (insufficient evidence), 
or \emph{Conflict} (contradictory 
evidence). We then present a methodology that exploits encoded information present in internal representations for prediction. 
Specifically, we train a lightweight \textit{triage router} (which is a logistic regression based prediction module)
on activations extracted from a single 
hidden layer of the language model, using 
a controlled diagnostic benchmark that 
isolates evidence-state decisions under 
matched inputs. We evaluate across 16 
transformer-based language models spanning 
90M to 32B parameters, diverse 
architectures, and training regimes. 
The router consistently and substantially 
outperforms prompt-based and 
RAG-specialised baselines across almost all 
models, achieving up to 0.91 accuracy 
and reducing false answer rates by up 
to 75\% relative to the strongest 
prompt-based competitor, with no 
additional tokens or inference cost.

Furthermore, we analyse \textit{where} evidence-state information is linearly decodable from the model's internal representations. Layer probing experiments show that the most informative representations consistently emerge in the middle layers of the network, a pattern that holds across model families and sizes. Hidden-state patching confirms that targeted intervention at the best probe layer reliably reshapes downstream model behaviour.

 Our core contributions in the paper are: (i) A three-way RAG triage formulation that explicitly distinguishes \emph{Answer}, \emph{Refuse}, and \emph{Conflict} evidence states, with a controlled diagnostic benchmark designed to allow us to evaluate the decisions; (ii) A lightweight triage router that 
    decodes evidence state from a single 
    hidden layer, outperforming all 
    prompt-based and RAG-specialised 
    baselines across 16 models without 
    additional inference cost; and (iii) An analysis of where evidence-state information is linearly decodable, together with hidden-state interventions that test its influence on downstream behaviour.

\section{Methodology}
In our experimentation, for a given question $q$ and a set of retrieved documents $D = {d_1, \dots, d_k}$, the system is expected to decide whether the retrieved evidence (i) answers a question, (ii) refuses to answer, or (iii) identifies conflicting answers within the provided context. We denote the triage label by

\begin{equation}
y \in \left\{\textsc{Answer}, \textsc{Refuse}, \textsc{Conflict}\right\}.
\end{equation}

Before the language model generates the answer, we propose a system to reliably predict $y$ to reduce instances where the model should refrain from answering or report conflicting cases. An instance is labelled \textsc{Answer} if the retrieved documents provide sufficient evidence and have only \textsc{one} grounded answer to the question. If the instance has \textsc{more than one} grounded answer to the question, it is labelled as \textsc{Conflict}. Finally, the instance is labelled \textsc{Refuse} if the retrieved documents do not contain enough information to answer the question. We can utilise signals of LMs as distinguishing features using a lightweight classifier that can be used to study the internal states and distinguish decisions.

\subsection{Dataset Construction}
Our objective is to isolate a specific decision problem: whether the provided evidence is sufficient to answer, insufficient, or conflicting. We therefore use and modify three popular datasets: TriviaQA \citep{Joshi2017TriviaQA}, HotpotQA \citep{Yang2018HotpotQA}, and Natural Questions \citep{Kwiatkowski2019NaturalQuestions}. This design focuses on interpretability by reducing parametric knowledge leakage and maintaining variability in the dataset. We randomly sample 797 question--answer--evidence triples from each source, resulting in 2,391 base questions in total. For each question, we keep the supporting document associated with the answer and treat it as the evidence document.

%To reduce the effect of the model answering a question using its parametric knowledge, motivated by the strategy introduced in \citep{Xie2024AdaptiveChameleon}, we systematically apply counterfactual entity substitution using Gemini~2.5~Pro. 
To reduce parametric knowledge leakage, we apply counterfactual entity substitution following the strategy of \citep{Xie2024AdaptiveChameleon}, implemented via  Gemini~2.5~Pro. We note that the substitution is applied globally and consistently: i.e., whenever an original entity appears across any document in the dataset, it is replaced with the same counterfactual, preventing models from recovering the original answer through cross-document consistency. Each substituted entity is unique within the dataset to prevent repetition-based inference.
%This entity substitution is applied consistently not only in the gold document, but also globally in any other documents in the dataset where the original entity appears. The remainder of the text and the styling template is unchanged apart from changing the particular selected entity. To reduce residual stylistic cues, we keep track of the replaced entity and do not repeat the same replaced entity in any other target answer.

To emulate a standard RAG setting, we retrieve candidate documents using BM25, a popular lexical retrieval baseline that is widely used in RAG pipelines \citep{Wang2024RAGBestPractices}. For each question, we retrieve documents from other contexts in the dataset, which can be gold contexts for other questions, and create three evidence configurations.

For each question, we construct three five-document context sets. The \textsc{Answer} set contains the gold document together with four BM25-retrieved distractor documents, with the gold document placed at a random position for all \textsc{Answer}, and \textsc{Conflict} cases. The \textsc{Refuse} set contains five BM25-retrieved non-gold documents. The \textsc{Conflict} set contains two gold documents, where the original gold document and a conflicting version created by replacing the answer entity with a different counterfactual conflicting entity using Gemini~2.5~Pro are combined with three distractor documents retrieved using BM25.

Each question gives us three aligned instances, one for each expected label. The final dataset contains $2{,}391 \times 3 = 7{,}173$ instances in total, with every instance containing exactly five documents. We use a question-level split of 70\% training, 10\% validation, and 20\% test data that avoids split leakage and remains constant across our experiments.

\paragraph{Quality control.}
We apply automated and human verification to ensure dataset quality. First, Gemini~2.5~Pro is used to 
verify that, outside of the gold documents, no distractor document contains the substituted answer entity. Second, an internal consistency check was performed by a co-author on a stratified sample of 10\%,  verifying entity replacement correctness and document coherence. We report an internal agreement rate of 100\% on this sample set.
%check the contexts so that, other than the gold documents, the other context documents do not contain the answer. Second, a human evaluator (a PhD student), who is also the co-author of this paper, manually verified the correctness of the Gemini automated procedure, including entity replacement in documents. 
The resulting benchmark is intentionally controlled and should be viewed as a diagnostic testbed rather than as a replacement for naturalistic RAG evaluation. Our aim is to situate the dataset to isolate whether models internally distinguish answerable, unanswerable, and conflicting evidence conditions under matched inputs. %To avoid parametric knowledge leakage, we claim the dataset to be a experimental testbed of RAG-style diagnostic test.

\paragraph{On Parametric Knowledge Leakage}
A core assumption of our benchmark is that entity substitution prevents models from answering questions from parametric memory alone. To verify this empirically, we measure each model's accuracy on the \textsc{Answer} instances when \textit{no context documents are provided} (\textbf{NO\_CTX} condition). Note here that if substitution is indeed effective, models should perform near chance ($\approx$0.33 for a three-way problem, or near 0.0 if measured only against the substituted answer string).

\subsection{Evaluation Metrics}

The overall classification accuracy over the three labels is our \textbf{Accuracy} metric. \textbf{Macro-F1} is the macro-averaged F1 across \textsc{Answer}, \textsc{Refuse}, and \textsc{Conflict}. Lastly, \textbf{False Answer Rate (FAR)} is our reliability metric, which is defined as the fraction of examples whose true label is either \textsc{Refuse} or \textsc{Conflict}, but are incorrectly predicted as \textsc{Answer}.
FAR captures the cases where the model generates an answer when it should not do so. Additionally, for evaluation, we perform a substring match to determine whether the language models produced a refusal/conflict string, validating the accuracy of the baselines. Furthermore, we report 95\% confidence intervals using 10,000 bootstrap resamples with replacement from the 1,435 test instances, recomputing metrics on fixed test predictions without retraining; intervals are the 2.5th--97.5th empirical percentiles, using a fixed random seed of 42.

\subsection{Baselines}
\label{sec:baselines}
We compare our router against prompt-based and RAG-specialised baselines: \textbf{ATTR}: The model is instructed to answer using only the retrieved documents with an explicit refusal or conflict policy. We construct and use the prompt, taking inspiration from the \textit{Attribute} style structure presented by \cite{Zhou2023ContextFaithful}. \textbf{KRE}: Based on \cite{ying2024intuitive}, we adapt their \textit{instruction-with-hint} prompting design and construct the baseline prompt. \textbf{OPIN\_INSTR}: We adapt the opinion based prompting framework of \cite{Zhou2023ContextFaithful} and construct the baseline. \textbf{FaithfulRAG style CoT Prompt:} Inspired by \cite{Zhang2025FaithfulRAG}, we adapt their CoT style prompting method and construct this baseline. \textbf{Chat-QA-1.5:} We utilise a specialised pre-trained RAG model introduced by \cite{liu2024chatqa}, with Llama3-8B model as its backbone. \textbf{Self-RAG:} We evaluate another specialised pre-trained RAG model introduced by \cite{asai2023selfrag}. The backbone of the model is the Llama2-7B model. We additionally evaluate two surface-text baselines without hidden states: \textbf{TF-IDF with logistic regression} over the concatenated question and documents, and a zero-shot \textbf{NLI DeBERTa cross-encoder}. Moreover, in addition to the above-mentioned baselines, we also test the language models on vanilla prompts and RAG setup without providing any context (\textbf{NO\_CTX}). These two baselines can be considered diagnostic lower bounds. 

\subsection{Primary Router: Activation as Feature}
\label{sec:router_act}

\textbf{Feature Extraction:} To train our classifier, instead of looking only at the output produced by the language model, we use a forward hook and extract features for every layer of the model for the last token of the prompt. We use the Naive Prompt of our baseline for feature extraction. Based on previous research, the last prompt token has shown characteristics of carrying information for the entire input sequence~\citep{AlainBengio2017Probing, Wendler2024ProbeMultilingual}.

For an input instance $x=(q,D,p)$ with documents $D$, and prompt template $p$, we extract features for every layer of the model from Hidden States ($H$), and the MLP Outputs ($M$) of each layer $l$ of the language model. 

\textbf{Training:} We train multinomial logistic regression models and perform a sweep over all layers ($l$). For each layer of the language model, we developed two distinct logistic regression models: one utilising $H$ features and the other utilising $M$ features. The specific layer and feature set (either $H$ or $M$) that achieved the highest classification accuracy on the validation set was selected for the final router ($r$).
\subsection{Secondary Router: Attention Weights as Feature}
\label{sec:router_att}
To compare with our first router, we create an additional router, a lightweight model based on attention weights that is trained as a classifier on features derived from the alignment between the retriever score distribution and the LM's attention over documents. This is motivated by previous work exploring attention weights as proxy signals for relevance and faithfulness \citep{tian2026reattn}.

\textbf{Feature Extraction:} We extract seven scalar features per layer from the last query token's attention row over retrieved document segments: the top-1 attention mass, top-2 attention mass, the gap between them, attention entropy, retriever score alignment, attention coverage, and the attention margin.

\textbf{Training:} Similar to our main router discussed in \ref{sec:router_act}, we train each layer on the validation data, and based on the highest accuracy, we select the particular layer for training on the training data. The training configuration is identical to our primary router.
\subsection{Language Models}
\label{sec:lm}
For our experiment, to ensure generalisation of our method, we chose 16 different transformer-based language models spanning different parameter sizes (90M to 32B), architectural designs, and training methods summarised in Table \ref{tab:models} in the Appendix.

\subsection{Implementation Details}
\label{sec:impl}
All experiments were conducted using NVIDIA A100 (80GB) GPUs hosted on a High-Performance Computing (HPC) cluster. For training our routers, we used Multinomial Logistic Regression along with StandardScaler with Sigmoid calibration. For all of our experiments, we use greedy decoding with a fixed maximum length of 64 tokens to maintain consistency across models. Only for the language model gpt-oss-20b, we keep an exception. As this is the only reasoning model in our test, and it generates reasoning traces and tokens, along with the actual answer, the maximum length of 1024 tokens was used. We additionally evaluate selective prediction at 80\% and 50\% coverage, with full results reported in (Appendix~\ref{app:calibration}).

\begin{table*}[t]
\centering
\small
\setlength{\tabcolsep}{3.5pt}
\caption{Comparison of accuracy with prompt-based, RAG-specialised, and surface-text baselines.
ATTR, KRE, FaithRAG, OPIN\_INSTR, Vanilla, and NO\_CTX are prompt-based
baselines. RA denotes our Router (Activation Features). For FAR, lower values
are better.}
\label{tab:main_results}

\resizebox{\textwidth}{!}{%
\begin{tabular}{lcccccc|ccc}
\toprule
& \multicolumn{6}{c|}{\textbf{Prompt-based Techniques}}
& \multicolumn{3}{c}{\textbf{Ours}} \\
\cmidrule(lr){2-7} \cmidrule(lr){8-10}

\textbf{Model}
& \shortstack{\textbf{ATTR}\\{\scriptsize [\citealp{Zhou2023ContextFaithful}]}}
& \shortstack{\textbf{KRE}\\{\scriptsize [\citealp{ying2024intuitive}]}}
& \shortstack{\textbf{FaithRAG}\\{\scriptsize [\citealp{Zhang2025FaithfulRAG}]}}
& \shortstack{\textbf{OPIN\_INSTR}\\{\scriptsize [\citealp{Zhou2023ContextFaithful}]}}
& \textbf{Vanilla}
& \textbf{NO\_CTX}
& \textbf{Acc.}
& \textbf{Macro-F1}
& \textbf{FAR} \\

\midrule
Falcon-H1-Tiny-90M      & 0.34 & 0.34 & 0.35 & 0.34 & 0.33 & 0.33 & 0.69 & 0.68 & 0.13 \\
Falcon3-7B-1.58bit      & 0.35 & 0.35 & 0.31 & 0.33 & 0.22 & 0.32 & 0.76 & 0.73 & 0.12 \\
OLMo-2-1B               & 0.31 & 0.36 & 0.33 & 0.33 & 0.29 & 0.33 & 0.75 & 0.73 & 0.14 \\
OLMo-2-1B-SFT           & 0.35 & 0.32 & 0.31 & 0.35 & 0.27 & 0.32 & 0.77 & 0.76 & 0.11 \\
OLMo-2-1B-DPO           & 0.35 & 0.32 & 0.31 & 0.35 & 0.27 & 0.32 & 0.77 & 0.76 & 0.11 \\
OLMo-2-1B-Instruct      & 0.36 & 0.33 & 0.34 & 0.33 & 0.28 & 0.33 & 0.78 & 0.78 & 0.12 \\
OLMo-3-7B-Instruct      & 0.34 & 0.34 & 0.36 & 0.34 & 0.34 & 0.33 & 0.81 & 0.81 & 0.13 \\
OLMo-3.1-32B-Instruct   & 0.64 & 0.64 & 0.67 & 0.54 & 0.44 & 0.33 & 0.90 & 0.90 & 0.06 \\
Qwen3.5-2B              & 0.33 & 0.34 & 0.33 & 0.34 & 0.33 & 0.33 & 0.85 & 0.80 & 0.08 \\
Qwen3-4B-Instruct       & 0.59 & 0.80 & 0.62 & 0.80 & 0.51 & 0.33 & 0.89 & 0.88 & 0.08 \\
Qwen3.5-9B              & 0.34 & 0.32 & 0.34 & 0.33 & 0.33 & 0.36 & 0.91 & 0.89 & 0.07 \\
Qwen3.5-27B             & 0.50 & 0.37 & 0.36 & 0.35 & 0.34 & 0.33 & 0.91 & 0.90 & 0.08 \\
Llama-3.2-3B            & 0.34 & 0.34 & 0.33 & 0.34 & 0.32 & 0.33 & 0.82 & 0.81 & 0.09 \\
Phi-mini-MoE-Instruct   & 0.42 & 0.42 & 0.37 & 0.44 & 0.34 & 0.33 & 0.83 & 0.82 & 0.11 \\
Granite-3.1-8B-Instruct & 0.49 & 0.50 & 0.55 & 0.39 & 0.34 & 0.33 & 0.88 & 0.87 & 0.07 \\
gpt-oss-20b             & 0.80 & 0.77 & 0.82 & 0.61 & 0.34 & 0.24 & 0.82 & 0.81 & 0.07 \\

\midrule
& \multicolumn{9}{c}{\textbf{Specialised and Surface-text Baselines}} \\
\cmidrule(lr){2-10}

& \multicolumn{2}{c}{\shortstack{\textbf{Self-RAG}\\
{\scriptsize [\citealp{asai2023selfrag}]}}}
& \multicolumn{2}{c}{\shortstack{\textbf{Chat-QA-1.5}\\
{\scriptsize [\citealp{liu2024chatqa}]}}}
& \multicolumn{2}{c}{\textbf{TF-IDF + LR}}
& \multicolumn{3}{c}{\textbf{NLI DeBERTa}} \\

\midrule

\textbf{Acc.}
& \multicolumn{2}{c}{0.340}
& \multicolumn{2}{c}{0.370}
& \multicolumn{2}{c}{0.356}
& \multicolumn{3}{c}{0.369} \\

\bottomrule
\end{tabular}%
}
\end{table*}

\section{Results}
\subsection{Model-wise Accuracy}

\begin{table*}[t]
\centering
\caption{Comparison of average accuracy of the best performing baselines (ATTR, KRE, FaithRAG, and OPIN\_INSTR) and our router-based model (RA) on the top three highest accuracy language models. The table also compares category-wise accuracy for each language model in correctly classifying Answer, Refuse, and Conflict states. For each metric, the best value is highlighted (Higher value is better for Acc. and Macro-F1. Lower value is better for FAR).}
\label{tab:baseline_vs_ours}
\footnotesize
\setlength{\tabcolsep}{4pt}
\renewcommand{\arraystretch}{1.1}

\resizebox{\textwidth}{!}{%
\begin{tabular}{lcccccc|cc|cc|cc}
\toprule
\multirow{3}{*}{Model}
& \multicolumn{6}{c|}{Overall}
& \multicolumn{2}{c|}{Answer}
& \multicolumn{2}{c|}{Refuse}
& \multicolumn{2}{c}{Conflict} \\
\cmidrule(r){2-7}
\cmidrule(r){8-9}
\cmidrule(r){10-11}
\cmidrule(l){12-13}
& \multicolumn{2}{c}{Acc.}
& \multicolumn{2}{c}{Macro-F1}
& \multicolumn{2}{c|}{FAR}
& \multicolumn{2}{c|}{}
& \multicolumn{2}{c|}{}
& \multicolumn{2}{c}{} \\
\cmidrule(r){2-3}
\cmidrule(r){4-5}
\cmidrule(r){6-7}
& Base & RA & Base & RA & Base & RA & Base & RA & Base & RA & Base & RA \\
\midrule
gpt-oss-20b
& 0.73 & \best{0.82}
& 0.75 & \best{0.81}
& 0.19 & \best{0.07}
& 0.66 & \best{0.69}
& \best{0.94} & 0.93
& 0.58 & \best{0.84} \\

Qwen3-4B-Instruct
& 0.70 & \best{0.88}
& 0.70 & \best{0.88}
& 0.28 & \best{0.08}
& 0.73 & \best{0.85}
& 0.66 & \best{0.87}
& 0.76 & \best{0.94} \\

Olmo-3.1-32B-Instruct
& 0.61 & \best{0.90}
& 0.70 & \best{0.90}
& 0.19 & \best{0.06}
& 0.41 & \best{0.86}
& \best{0.97} & 0.93
& 0.44 & \best{0.90} \\
\bottomrule
\end{tabular}%
}
\end{table*}

Our primary findings are presented in Table \ref{tab:main_results}. Across all 16 models, our activation-feature router (RA in the table) consistently outperforms all prompt-based and RAG-specialised baselines on accuracy (we also report our Macro-F1, and FAR). We additionally evaluated specialised RAG models, \textbf{Self-RAG} \citep{asai2023selfrag} and \textbf{Chat-QA-1.5} \citep{liu2024chatqa}, which gained overall accuracies of 0.34 and 0.37, respectively, indicating near-chance performance. The surface-text baselines also remain near chance: \textbf{TF-IDF+LR} achieves 0.356 accuracy and 0.352 Macro-F1, while zero-shot \textbf{NLI DeBERTa} achieves 0.369 accuracy and 0.338 Macro-F1.

We highlight an important pattern where most prompt-based baselines produce accuracy near chance (0.33) for the majority of tested models. To further understand why, we analyse the failure cases and we observe that smaller and non-instruction-tuned models tend to \textit{always answer}, ignoring refusal or conflict instructions entirely which indicates that the instruction following capability required to produce structured refusal behaviour is simply absent. Second, larger instruction-tuned models that do follow refusal instructions tend to \textit{over-refuse}: they identify insufficient evidence correctly but also refuse on \textsc{Answer} instances, collapsing into a two-class behaviour that inflates \textsc{Refuse} predictions at the expense of \textsc{Answer} and \textsc{Conflict}. This over-refusal pattern is consistent with findings in prior work \citep{rottger2023xstest}. 

We present further evidence of this in Table~\ref{tab:baseline_vs_ours}, considering the top-performing models in terms of overall accuracy. We further note that neither failure mode is fixable by prompt engineering alone, as they reflect model-level behavioural biases rather than prompt interpretation errors. This further motivates our approach, where we are able to obtain information directly from the internal activations before generation begins. We note that two broad patterns show up consistently.

%In the table, the accuracy of our routers and the compared baselines in distinguishing between answer, refuse, and conflict are presented. For all of the cases, our activation-feature-based router trained router outperforms the compared baselines. Besides this method gains better  performance, and utilises no additional tokens compared to prompt based baselines that we have.

%In Table \ref{tab:baseline_vs_ours}, we present the top-performing models in terms of overall accuracy and compare their performance with our method. Although gpt-oss-20b achieves the best overall accuracy among the tested language models, an exception was made for this model by setting a maximum output of 1024 (16x the maximum generation output of other models), as it is a reasoning model that generates reasoning traces along with the answer. Overall, our method outperforms all prompt-based methods and suggests performance superior to expensive reasoning-based models. It is evident from the table that out of the three models, the accuracy of correctly identifying refusal by the language models are prominent, and gpt-oss-20b, and OLMo-3.1-32B-Instruct beats our method. This finding can be related to the previous researches where researchers have identified potential issues of language models in over-refusing \citep{rottger2023xstest}. But in terms of identifying conflicting documents, it can be seen that language models get confused when conflicting cases are present, and in those cases, our router based method outperforms in identifying the conflicting cases correctly. 

We also highlight the NO\_CTX accuracy for all 16 models. Across the full model suite, NO\_CTX accuracy ranges from 0.24 to 0.36, confirming that  models cannot reliably recover the substituted answers  from parametric knowledge alone. Notably, even the largest models tested (OLMo-3.1-32B-Instruct,  gpt-oss-20b) do not exceed 0.33 and 0.24 respectively in the NO\_CTX condition, suggesting that entity substitution is effective across model scales. We therefore treat the NO\_CTX condition as a leakage audit.

\subsection{Layer Probe}
\begin{figure}[h] \centering \includegraphics[width=1\textwidth]{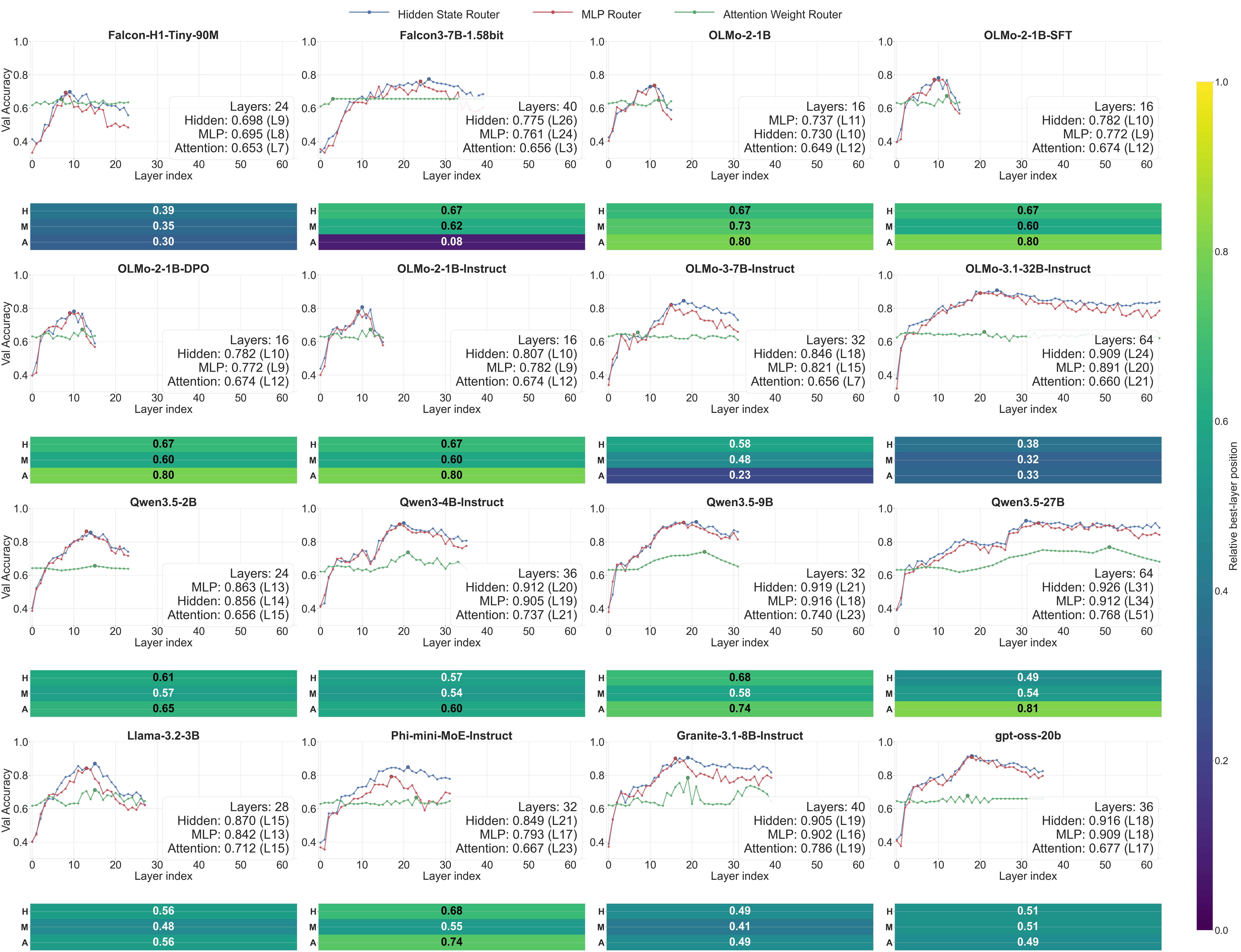} \caption{Layer-wise validation accuracy of three routing strategies across sixteen language models. Each subplot shows validation accuracy (y-axis) as a function of transformer layer index (x-axis) for three router types: Hidden State Router (blue), MLP Router (red), and Attention Weight Router (green). Filled circles mark the best-performing layer per router. The heatmap strip below each plot shows the relative best-layer position (best layer / total layers), where darker colours indicate earlier layers and brighter colours indicate later layers. H, M, and A denote Hidden State, MLP, and Attention routers, respectively. The annotation box reports the total layer count and peak validation accuracy with its layer index for each router. An enlarged version of this figure is provided in Figure~\ref{fig:zoomed_acc}, Appendix~\ref{app:zoom}.} \label{fig:results_graph} \end{figure}

For each model, we train each layer using hidden-state, MLP, and attention-weight features. We visualise and compare all the layers of the language models and the validation accuracy for each layer in Figure \ref{fig:results_graph}.

The heatmap portion in Figure \ref{fig:results_graph} demonstrates the depth and region within the language model, based on the layer on which the best layer based on the extracted features exists for each model. For most of the models, we can see peak dominance mostly in the middle region. Although hybrid and quantised structured models, Falcon-H1-Tiny-90M and Falcon3-7B-1.58bit, show different characteristics in terms of best-layer performance, based on the extracted feature.

\section{Mechanistic Analysis}

\subsection{Router-type hierarchy: hidden state is usually best}
A consistent result is that hidden-state features are the strongest in almost all the models, followed by MLP feature-based routers and attention-weight-feature-based ones. Since hidden states reflect the integrated residual-stream representation after passing through the attention and MLP blocks, they are expected to provide more features for our routers. MLP outputs reflect only the feedforward branch, and attention weights primarily indicate where the model attends rather than the full content of what has been computed. As a result, hidden states in most cases provide rich and transferable routing signals.

\subsubsection{The OLMo-2-1B base exception}
For OLMo-2-1B base, the MLP router slightly outperforms the hidden-state router. However, the behaviour reverses in the SFT, DPO, and Instruct variants. This indicates that although in base variants the MLP router has rich distinguishing features, after post-training, it becomes more integrated into the hidden states without altering the depth region.

\subsection{Relative layer position: where models peak}
Overall, among the 16 evaluated language models, most of the feature-extracted models show an inverted-U trajectory that peaks in the middle-to-upper-middle portion of the network. The relative position of the peak remains broadly similar, irrespective of the size of the language models.

\subsubsection{Falcon-H1-Tiny-90M and OLMo-3.1-32B-Instruct peak unusually early}
A distinct result in the heatmap portion of the Figure \ref{fig:results_graph} is that Falcon-H1-Tiny-90M, the smallest of the language models tested, peaks earlier than all other models for all three router types. Because of its small structure and hybrid Mamba/attention design, the behaviour is different compared to standard dense transformers.

For OLMo-3.1-32B-Instruct, it is evident that task-specific information is linearly decodable in the early layers as well. The instruction-tuned OLMo model is initialised from an OLMo-3-Think-SFT checkpoint and is then optimised for non-reasoning chat, multi-turn preference alignment, and function calling \citep{teamolmo2025olmo3}. This optimisation could be a factor to make the model linearly accessible in the early layers, and can be hypothesised based on our findings. 

\subsection{Model-specific sources of layerwise variation}

The OLMo-2 family shows the most stable pattern across variants. The base, SFT, DPO, and Instruct models have nearly identical curve shapes and peak in a similar layer range, with fine-tuning mainly increasing peak accuracy rather than moving the optimal layer. This suggests that the backbone architecture, more than the post-training stage, determines where the most informative representations lie.

Falcon3-7B-1.58bit is unusual in that its attention router peaks extremely early, while its hidden-state and MLP routers retain a more standard middle-layer optimum. A plausible explanation is that low-bit quantisation affects attention-derived signals more strongly than activation-derived signals, making attention weights less stable across depth.

\subsubsection{Granite-3.1-8B-Instruct: why the curve is more irregular}

Granite-3.1-8B-Instruct exhibits the strongest layer-to-layer fluctuation among the dense models. Several architectural features likely contribute to this behaviour. First, Granite uses explicit scaling terms such as \texttt{residual\_multiplier}, \texttt{attention\_multiplier}, \texttt{embedding\_multiplier}, and \texttt{logits\_scaling}, which suggest a more tightly regulated signal flow through the network.

Second, Granite is one of the deepest models in the comparison, which increases the chance that representations are repeatedly re-encoded rather than smoothly refined. Third, it is the only model here with non-zero attention dropout. Finally, its grouped-query attention may reduce the distinctiveness of raw attention outputs relative to the hidden state. Together, these properties likely explain why Granite shows a more jagged and inconsistent layer-wise profile. A comparison of the architectures of the tested models is presented in Figure \ref{fig:architecture} in the Appendix.

\section{Ablation}
We evaluate the robustness of our router design through cross-model transfer, prompt robustness, hidden state interventions, and additional diagnostic
ablations.

\begin{comment}
\begin{figure}[htbp]
    \centering
    \includegraphics[width=1\textwidth]{causal.png}
    \caption{Causal Patching Success}
    \label{fig:results_causal}
\end{figure}
\subsection{Causal Patching}
For our router-based models, we test whether a specific internal representation is functionally responsible for a downstream prediction. In this method, we replace a target activation with a source activation and measure whether the model's prediction changes accordingly.

Our causal patching experiment is inspired by the activation patching works introduced by \cite{geiger2021causal} and \cite{meng2022locating}. We conduct this experiment on our validation dataset on 40 random pairs of answer-refuse and answer-conflict cases.

Figure \ref{fig:results_causal} represents that causal patching is successful across all models and patch directions, with flip-to-source rates always being above the chance baseline of 0.33. Performance is consistently strong for refuse-to-answer and conflict-to-answer cases, often exceeding 0.90 in instruction-tuned models. Although answer-to-refuse is the weakest, it still remains above chance. These results indicate that the router-layer representation is causally influential.
\end{comment}
\subsection{Cross-Model Transferability}
\begin{wrapfigure}{l}{0.43\textwidth}
    \vspace{-8pt}
    \centering
    \includegraphics[width=0.33\textwidth]{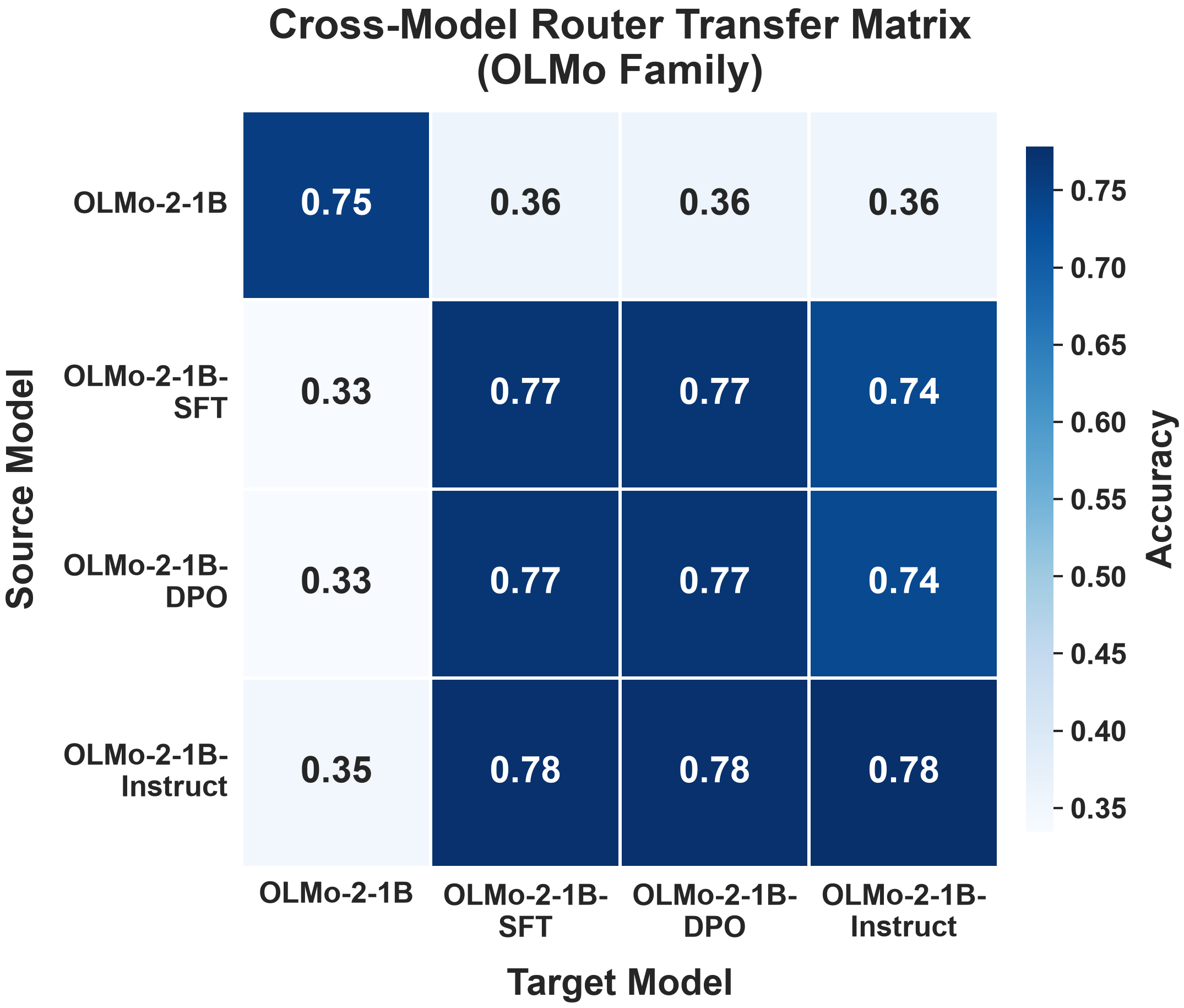}
    \caption{OLMo Model Transferability.}
    \label{fig:results_olmo}
    \vspace{-10pt}
\end{wrapfigure}
To conduct cross-model transfer within the OLMo-2 family (OLMo-2-1B, OLMo-2-1B-SFT, OLMo-2-1B-DPO, and OLMo-2-1B-Instruct), we train a router on one model and test it on the other corresponding models. Figure \ref{fig:results_olmo} indicates the results. Routers trained on the fine-tuned variants SFT, DPO, and Instruct transfer well across each other, indicating that these model variants preserve a similar geometry for our task classification. In contrast, transfer to and from the base variant does not generalise, showing that fine-tuning introduces a linear representation that the base variants lack.
\subsection{Prompt Robustness}
To evaluate the robustness of our system on different prompts, we train our router on one specific prompt and test the accuracy on different prompts on a subsample of validation data. The details of the prompt, along with the produced heatmap, are provided in Appendix \ref{sec:prompt}. By analysing Figure~\ref{fig:results_prompt}, we can see that the performance is largely stable across Prompts P0 to P3, where minimal lexical variations are applied. The performance drops on Prompt P4, which uses question-first ordering, suggesting that the system is sensitive to the position of the questions within the context of the prompt. When a style wrapper (Prompt P5) is applied, it affects smaller base models more than larger ones.

\subsection{Hidden-state Patching}
%To validate our use of the layer with the highest validation accuracy for router training and evaluation, we perform a hidden-state patching experiment inspired by \cite{meng2022locating}. We test whether the selected router layer is functionally important within the transformer by constructing source-target pairs from different triage modes (ANSWER, CONFLICT, and REFUSE). For each pair, we run the language model on the target example before and after replacing its hidden state at the selected layer with the corresponding hidden state from the source example.

%%% try: 
To validate that the router layer is a functionally active control point, we perform a hidden-state patching experiment following \cite{meng2022locating}. For each of 100 source–target pairs drawn from different triage classes (ANSWER, CONFLICT, and REFUSE), we replace the target example's hidden state at the selected router layer with the corresponding hidden state from the source example, then measure how downstream router predictions shift. The aim of this is to establish that the selected layer is a fair control probe such that intervening at this layer is sufficient to reshape the downstream prediction. This provides evidence that the selected layer acts as a functional control
point for downstream routing behaviour. 

We present the results in Figure~\ref{fig:causal_combined} across 11 language models. Patching the best probe layer produces a substantial increase in downstream source-class prediction rates across all patch directions and models. The effect is strongest for refuse$\rightarrow$answer and conflict$\rightarrow$answer directions, consistently exceeding 0.70 in instruction-tuned models. The answer$\rightarrow$refuse direction produces smaller but reliable shifts, above chance in all tested models. These results confirm that the selected middle layer is directly relevant to the triage decision.

We additionally test whether the intervention propagates to free-form
generation, rather than only changing the downstream router prediction. Results show that the identified hidden-state signal can influence generated text, but does not provide a complete explanation of the model's generation process (results presented in Appendix~\ref{app:generation_patching}).

\subsection{Binary vs.\ Three-Way Formulation}
\label{sec:binary_ablation}
We compare the three-way router with a simpler binary version that does not treat CONFLICT as a separate class. The two settings achieve similar overall accuracy (0.829 vs.\ 0.832), but the three-way router gives better Macro-F1 (0.829 vs.\ 0.811) and a lower FAR (0.104 vs.\ 0.127). This suggests that keeping CONFLICT as a separate class helps reduce false answers without hurting overall performance. Full model-wise results are provided in Appendix~\ref{app:binary_ablation}. In addition, as a control, we train the probe with randomised labels. Accuracy drops to $0.332\pm0.013$ on average, close to chance; full results are reported in Appendix~\ref{app:random_labels}.

\subsection{Long-Context Feature-Source Ablation}
We compare the final prompt-token representation with two simple pooling
strategies: averaging the final content-token representation from each
document (Doc-end mean), and averaging the representations of all document
content tokens (Doc-token mean) on Qwen3-4B-Instruct and Llama-3.2-3B. Performance generally
drops as the context becomes longer, while the final-token representation
remains stronger than the pooling alternatives. Results are presented in Table~\ref{tab:long_context} of the Appendix.

Overall, the final-token representation remains stronger than the two simple pooling alternatives, but its performance declines as the number of retrieved documents increases. The drop is larger when a router trained at $k=5$ is
tested on longer contexts, indicating sensitivity to context-length distribution shift.

\subsection{Natural-Domain Transfer}

We additionally evaluate zero-shot transfer to RAGTruth \citep{Niu2024RAGTruth}, where routers trained without any RAGTruth supervision achieve 0.84--0.92 accuracy across four model families; full results and limitations are reported in Appendix~\ref{app:ragtruth}.
\section{Related Work}
\label{related}
Recent work on improving RAG faithfulness has primarily focused on different prompt-based techniques and specialised RAG models \citep{Zhou2023ContextFaithful,ying2024intuitive,Zhang2025FaithfulRAG,asai2023selfrag,liu2024chatqa}. While these methods are effective in some cases, this comes with additional token, reasoning, and text generation costs. Moreover, prompt optimisation does not necessarily generalise across language models. 

Research by \cite{Kadavath2022KnowWhatTheyKnow} suggests that model outcomes can be predicted internally. Extending on this, recent interpretability work also shows that hidden states can encode useful decision signals \citep{huang2025traceable, zhou2025retrieval, liu2025how, wang2025unveiling}. Researchers have shown how these signals can be utilised to probe correctness during reasoning, detect hallucinations from single responses, and study how retrieved and parametric knowledge interact inside RAG systems \citep{AlainBengio2017Probing,zhang2025reasoning,sriramanan2024llmcheck,redeep, zhou2025retrieval}. 

Our work is also related to recent approaches that use internal model states for RAG decisions. Probing-RAG \citep{proprag} uses intermediate hidden states for a binary
retrieve-more decision during generation, while SeaKR \citep{seakr} uses internal
uncertainty to guide retrieval and reasoning. RedeEP \citep{redeep} instead detects
hallucinations after generation by analysing parametric and retrieved-context
contributions. LLM Microscope \citep{mic} studies incorrect context using a single
perturbed document, whereas our CONFLICT setting captures disagreement
between retrieved sources. In contrast, our router makes a three-way
decision before generation begins, with CONFLICT treated as a separate
prediction target.

In addition to prompting and verifier-based approaches, we extend interpretability research and study whether the RAG evidence state itself \textsc{Answer}, \textsc{Refuse}, or \textsc{Conflict} is linearly decodable from internal activations before generation. Previous literature has not really addressed the three-way problem through layer probing and mechanistic exploration. Besides, work on activation steering \citep{steering, subra} suggests that vector steering can be a potential research direction for studying language models; we expand on it to explore RAG faithfulness. We conducted an ablation in applying Contrastive Activation Steering in Appendix \ref{sec:steer}, and aim to conduct future research on it.

\section{Conclusion}

In this paper, our focus was to examine whether language models internally represent the evidential state of their retrieved context before generating a response. Our experiments suggest that this is indeed possible and that the internal representations contain the necessary information. A simple logistic regression-based linear model trained on activations from a single hidden layer is reliably capable of distinguishing whether retrieved evidence is sufficient to answer, insufficient, or contradictory, across 16 language models spanning diverse architectures, sizes, and training regimens. We showcase that prompt-based approaches to RAG usually fail mostly because of model-level behavioural biases, suggesting that prompting alone is perhaps suboptimal. We observe that smaller models tend to always answer, larger models tend to over-refuse, and neither of these is capable of handling conflicting evidence. Our approach, on the other hand, bypasses these limitations entirely by operating on internal representations before generation begins, achieving up to 0.91 accuracy and reducing false answer rates by up to 75\% relative to the strongest prompt-based baseline with no additional token budget, while remaining extremely computationally efficient.

We also provide mechanistic findings, where we show that the most informative representations for the triage decision emerge consistently in the middle layers of the network, a pattern that holds across model families and is stable under post-training. Hidden-state patching confirms these layers are important control points and targeted intervention there reliably reshapes downstream behaviour. Among the three evidence states, \emph{Conflict} is the most practically dangerous as models generate confidently from contradictory documents, and still our approach shows that it is linearly decodable internally at the right layer. 

\section*{Acknowledgements}
This work was partially supported by the Mike Hudson Foundation, a non-profit AI donor fund, the City St George's, University of London, and
was supported in part by the Alan Turing Institute under Fundamental Research Project No. PP00029.

% In a RAG setup, it is important for language models to know and identify whether it should answer, refuse, or report conflicting cases situated within the context before generating an answer. Across the 16 language models that we tested, spanning scale, architecture, and post-training structure, a linear router trained on internal activations consistently outperforms prompt-only baselines and specialised RAG models on the three-way evidence-state task.

% Our method outperforms instruction-tuned models, even on base models, and achieves performance comparable to reasoning models with no additional cost from text generation tokens or prompt optimisation.

% There are limitations to our study. First, the benchmark is intentionally controlled, although in Appendix \ref{sec:arti}, we show that our method is robust to particular cues and artifacts. However, the experiment can be expanded further. Second, our router is backed by a linear model and is trained on the best-layer features. We conducted an ablation on this by experimenting on diverse features, and training the router using non-linear model in the Appendix \ref{sec:router}. 

% In summary, beyond the empirical gains, our findings provide evidence that transformer middle layers encode compact features across a variety of transformer models, and we believe that these states can be utilised to decode RAG systems in a variety of tasks.

\bibliography{colm2026_conference}
\bibliographystyle{colm2026_conference}

\clearpage

\appendix

\section{Artefact Robustness}
\label{sec:arti}
To test the robustness of our router and assess its reliability on formatting artefacts in the retrieved documents, we sample 20 examples from each of the ANSWER, REFUSE, and CONFLICT classes, for a total of 60 instances. We use automatic back-translation through a non-English language as a pivot using MarianMT/OPUS-MT and translate it back to English, which changes the structure of the retrieved document while preserving the semantics. We test our router, trained on the original data, on this rewritten set of retrieved documents using two language models. The router maintains an accuracy of \textbf{0.83} and \textbf{0.80} for Qwen3-4B-Instruct and Granite-3.1-8B-Instruct, respectively, suggesting that the trained router is independent of lexical and stylistic cues.

\section{Router-Design}
\label{sec:router}
\begin{table}[H]
\centering
\caption{Validation accuracy comparison between the baseline setup and ablation variants across three models. H means hidden state features, M means MLP block outputs from transformer. Top N suggests selecting the top-performing layers for training and testing the router. All-H/M suggests combining all the layers for training and testing the router.}
\label{tab:ablation_results}
\begin{tabular}{lccc}
\hline
\textbf{Setting} & \textbf{Granite-3.1-8B-Instruct} & \textbf{OLMo-2-1B} & \textbf{Qwen3-4B-Instruct} \\
\hline
Previous Val Acc (baseline) & 0.90 & 0.73 & 0.91 \\
\hline
\multicolumn{4}{l}{\textit{Layer aggregation}} \\
Top3-H   & 0.91 & 0.80 & 0.89 \\
Top3-M   & 0.90 & 0.81 & 0.91 \\
All-H    & 0.90 & 0.78 & 0.91 \\
All-M    & 0.90 & 0.80 & 0.90 \\
\hline
\multicolumn{4}{l}{\textit{Feature combination}} \\
Top1-(H+M) & 0.90 & 0.76 & 0.91 \\
\hline
\multicolumn{4}{l}{\textit{Classifier capacity}} \\
MLP-Top1-H & 0.91 & 0.83 & 0.92 \\
MLP-Top1-M & 0.90 & 0.80 & 0.91 \\
\hline
\end{tabular}
\end{table}
The results presented in Table~\ref{tab:ablation_results} support our claim that the routing signal is linearly decodable from a selected layer. We evaluate three language models under different architectural and design choices. Aggregating the top few layers or all layers yields only small and inconsistent gains relative to the best single-layer setup, especially for Granite-3.1-8B-Instruct and Qwen3-4B-Instruct, while OLMo-2-1B benefits more from broader aggregation. Combining hidden-state and MLP features also provides limited improvement. Furthermore, we evaluate a non-linear router based on a multilayer perceptron (MLP), and find that the gain is most pronounced for OLMo-2-1B compared to the other models. This suggests that smaller base models may benefit more from a slightly more expressive classifier, possibly because their routing signal is less directly linearly separable across transformer layers.
\section{Activation Steering}
\label{sec:steer}
The triage representation by probing can be represented by moving the model along the representation, which should shift its response. We test this state, which can be usable for control, using Contrastive Activation Addition, inspired by the work of \cite{steering}

\begin{figure}[H]
    \centering
    \includegraphics[width=1\textwidth]{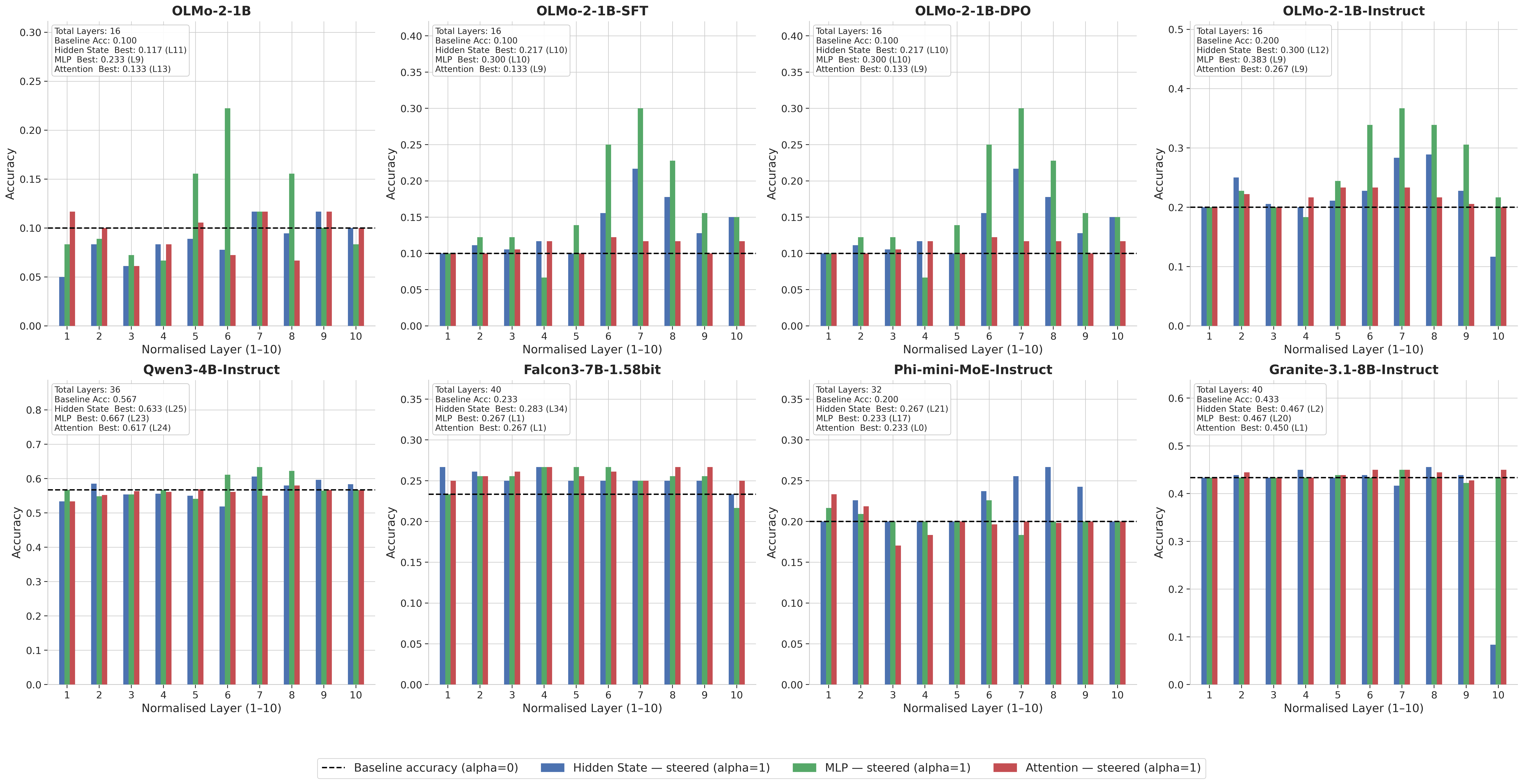}
    \caption{Mean accuracy across language models on a subset of 80 examples (40 Refuse, 40 Conflict) under vector steering on hidden state, MLP output, and attention output. The dashed horizontal line indicates the baseline accuracy. Layers are downsampled to 10 layers for ease of comparison.}
    \label{fig:caa}
\end{figure}

Figure \ref{fig:caa} presents our findings on steering activation. For each model, we compute steering vectors utilising a mean-difference estimator between positive and negative examples at each layer. We construct two vector types, REFUSE-vs-ANSWER and CONFLICT-vs-ANSWER, so that each direction represents how the internal state differs.

We compute vectors at three intervention sites: the residual stream, the MLP output, and the attention output. During generation, for layer \(l\), site \(s\), and steering strength \(\alpha\), we add the steering vector to the activation as
\[
\tilde{h}_l^{(s)} = h_l^{(s)} + \alpha v_l^{(s)}.
\]

We use \(\alpha \in \{0,1\}\), where \(\alpha=0\) is the unsteered baseline and \(\alpha=1\) is the steered setting, and sweep all layers of each model on a validation subset of 40 refuse and 40 conflict examples.

Baseline prompting and steering did not affect Falcon-H1-Tiny-90M and Llama-3.2-3B, as these models always generated an answer instead of producing refusal or conflict behaviour because of not being instruction-tuned, so we keep these out of our analysis. Figure \ref{fig:caa} shows that steering can partially control language model behaviours. Similar to our finding in the layer probe experiment, the strongest effects usually appear in middle-to-late layers. In contrast, in probing, hidden states were typically most decodable, but in steering, the gains mainly come from MLP features. On the other hand, attention steering is generally weaker and less consistent throughout all the tested language models.

The variation of gains after steering also varies by model and architecture. In OLMo, post-training increases steerability without changing the depth of the layer position, suggesting that it strengthens an existing feature rather than relocating it. In instruction-tuned models like Qwen and Granite, gains are smaller, likely because the baseline prompt already implemented stronger refusal/conflict behaviour. Very late-layer interventions can also hurt performance, especially in Granite, indicating that steering at the final layers can degrade performance badly. Overall, these results suggest that language models in RAG setups are both decodable and partly controllable.

\section{Prompt Templates \& Robustness}
\label{sec:prompt}
\begin{figure}[H]
    \centering
    \includegraphics[width=1\textwidth]{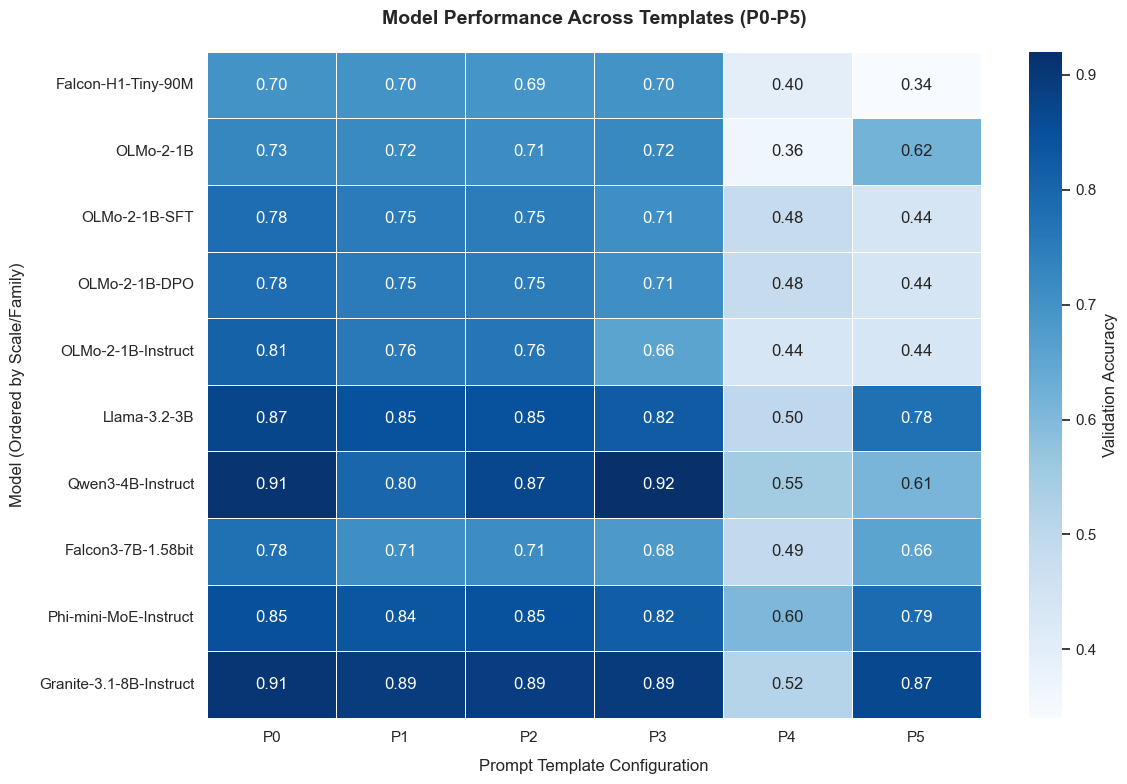}
    \caption{Prompt Robustness Heatmap}
\label{fig:results_prompt}
\end{figure}

Figure~\ref{fig:results_prompt} provides the full prompt-level result. Performance remains relatively stable across P0--P3, where the changes are mainly lexical, while a larger drop is observed for P4 with question-first ordering. P5 also affects some of the smaller models more strongly. This suggests that the router is reasonably robust to small prompt variations, but can still be affected by larger changes in prompt structure.

The prompt set consists of six variants, \texttt{P0} to \texttt{P5}, where each \texttt{P} denotes a different prompt template. The complete baseline prompts and deterministic output-parsing configuration are provided with the released code.

\begin{itemize}
    \item \textbf{P0 (Baseline):}
\begin{verbatim}
You are a helpful assistant. Use only the provided documents to answer.

[Document 1]:


Question: 
Answer:
\end{verbatim}

    \item \textbf{P1 (Instruction Paraphrase):}

\begin{verbatim}
Answer using only the provided documents. Do not use outside knowledge.

[Document 1]:


Question: 
Answer:
\end{verbatim}

    \item \textbf{P2 (Boundary Change):} 
\begin{verbatim}
You are a helpful assistant. Use only the provided documents to answer.

Passage 1:


Question:
Answer:
\end{verbatim}

    \item \textbf{P3 (No System Grounding):}
\begin{verbatim}
[Document 1]:


Question:
Answer:
\end{verbatim}

    \item \textbf{P4 (Question-First):}
\begin{verbatim}
You are a helpful assistant. Use only the provided documents to answer.

Question:

[Document 1]:


Answer:
\end{verbatim}

    \item \textbf{P5 (Structured Format):}
\begin{verbatim}
You are a helpful assistant. Use only the provided documents to answer.

<doc id="1">
</doc>

<question>...</question>
<answer>
\end{verbatim}
\end{itemize}

\section{Model Details}
Table~\ref{tab:models} summarises the 16 language models used in our experiments, covering different model families, parameter sizes, training stages, and architectures.

\begin{table}[H]
\centering
\caption{Model Suite Overview}
\label{tab:models}
\begin{tabular}{lllll}
\toprule
\textbf{Model Family} & \textbf{Variant} & \textbf{Parameters} & \textbf{Type} & \textbf{Architecture} \\
\midrule
Falcon  & Falcon-H1-Tiny-90M        & 90M  & Instruct & Hybrid      \\
OLMo    & OLMo-2-1B                 & 1B   & Base     & Dense       \\
OLMo    & OLMo-2-1B-SFT             & 1B   & SFT      & Dense       \\
OLMo    & OLMo-2-1B-DPO             & 1B   & DPO      & Dense       \\
OLMo    & OLMo-2-1B-Instruct        & 1B   & Instruct & Dense       \\
Qwen    & Qwen3.5-2B                & 2B   & Base     & Dense       \\
Llama   & Llama-3.2-3B              & 3B   & Base     & Dense       \\
Qwen    & Qwen3-4B-Instruct         & 4B   & Instruct & Dense       \\
Falcon  & Falcon3-7B-1.58bit        & 7B   & Instruct & quantised   \\
OLMo    & OLMo-3-7B-Instruct        & 7B   & Instruct & Dense       \\
Phi     & Phi-mini-MoE-Instruct     & 7.6B & Instruct & Sparse MoE  \\
Granite & Granite-3.1-8B-Instruct   & 8B   & Instruct & Dense       \\
Qwen    & Qwen3.5-9B                & 9B   & Base     & Dense       \\
GPT-OSS & gpt-oss-20b               & 20B  & Reasoning     & Dense         \\
Qwen    & Qwen3.5-27B               & 27B  & Base     & Dense       \\
OLMo    & OLMo-3.1-32B-Instruct     & 32B  & Instruct & Dense       \\
\bottomrule
\end{tabular}
\end{table}

\section{Hidden-State Patching}

\begin{figure}[H]
    \centering
    \includegraphics[width=\textwidth]{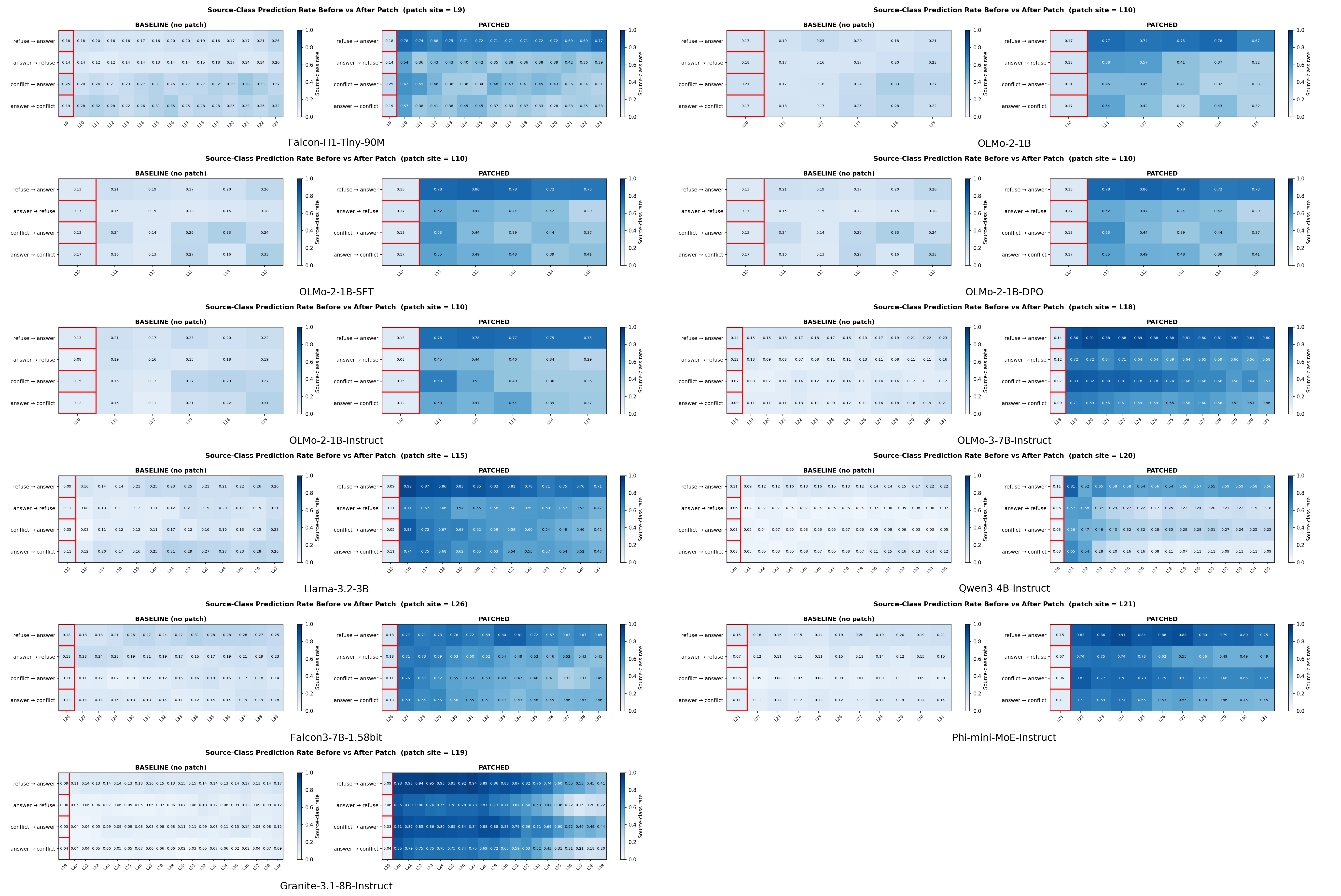}
    \caption{Patching on different tested language models. Label on y-axis, such as \textit{refuse $\rightarrow$ answer} means that a hidden state from a \textit{refuse} example is inserted into an \textit{answer} example at the patch site. The x-axis shows the patch layer and all subsequent layers, while each cell reports the proportion of examples whose downstream router prediction matches the source class. Higher values after patching indicate that the selected layer acts as a causal control point.}
    \label{fig:causal_combined}
\end{figure}

\section{Zero-Shot Transfer to RAGTruth}
\label{app:ragtruth}

To examine whether the routing signal transfers beyond our controlled benchmark, we evaluate the trained routers on RAGTruth \citep{Niu2024RAGTruth} without any additional training or adaptation on RAGTruth. We use the MS MARCO-derived portion of RAGTruth and evaluate four model families. The results are shown in Table~\ref{tab:ragtruth-transfer}.

\begin{table}[th]
    \centering
    \small
    \begin{tabular}{lcc}
        \toprule
        \textbf{Model} & \textbf{Accuracy} & \textbf{Macro-F1} \\
        \midrule
        Falcon3-7B-1.58bit       & 0.91 & 0.63 \\
        Granite-3.1-8B-Instruct  & 0.92 & 0.65 \\
        Llama-3.2-3B             & 0.84 & 0.65 \\
        Qwen3-4B-Instruct        & 0.91 & 0.63 \\
        \bottomrule
    \end{tabular}
    \caption{Zero-shot transfer to RAGTruth. The routers are evaluated without any RAGTruth training signal or in-domain adaptation.}
    \label{tab:ragtruth-transfer}
\end{table}

All four routers retain strong accuracy under this transfer setting, with accuracy ranging from 0.84 to 0.92. This provides evidence that the learned routing signal is not limited to the controlled construction used in our main experiments and can transfer to naturally occurring RAG examples.

There is an important limitation to this evaluation. First, RAGTruth does not provide a separate annotation for conflicting retrieved evidence. We therefore evaluate transfer in a binary setting, and this experiment does not test transfer of the three-way ANSWER--REFUSE--CONFLICT formulation directly.
We treat the RAGTruth experiment as evidence of natural-domain transfer rather than as a replacement for the controlled three-way evaluation. A fully annotated naturalistic benchmark containing answerable, insufficient, and conflicting evidence would allow a more direct test of three-way transfer.

\section{Additional Results}
\begin{figure}[H]
    \centering
    \includegraphics[width=1\textwidth]{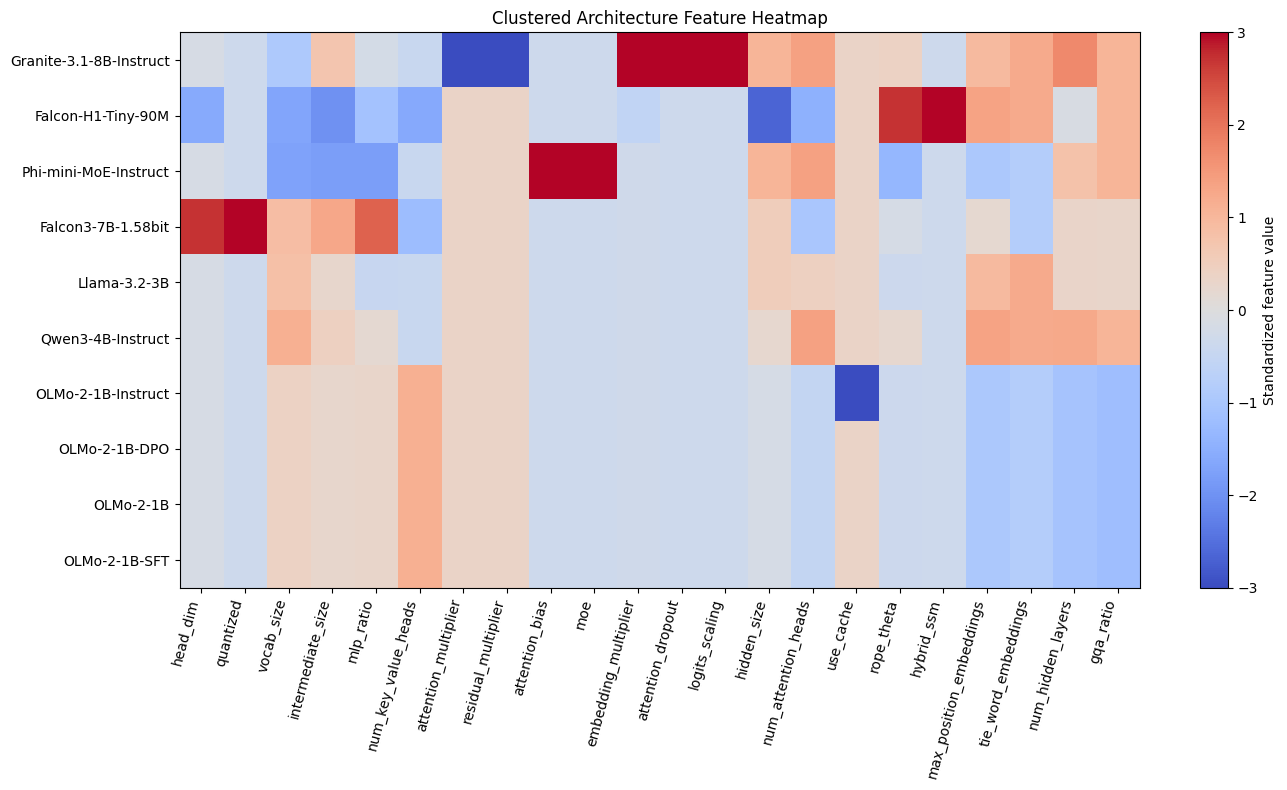}
    \caption{Architectural Feature Heatmap}
    \label{fig:architecture}
\end{figure}

Figure~\ref{fig:architecture} provides an architectural comparison of the models used in the main experiments. The figure provides additional context for the model-specific layer-wise differences.

\section{Calibration and Selective Abstention}
\label{app:calibration}

We evaluate whether calibrated router confidence can support selective
abstention by retaining only the most confident predictions. Table~\ref{tab:selective_abstention}
reports Accuracy and False Answer Rate (FAR) at 80\%, and 50\%
coverage. On average, reducing coverage from 100\% to 80\% increases
accuracy while reducing FAR. At 50\% coverage, average accuracy reaches 0.943 and FAR decreases to
0.013.

\begin{table}[th]
\centering
\small
\setlength{\tabcolsep}{3pt}

\resizebox{0.78\linewidth}{!}{%
\begin{tabular}{lcccc}
\toprule
\textbf{Model} &
\textbf{Acc.@80\%} &
\textbf{FAR@80\%} &
\textbf{Acc.@50\%} &
\textbf{FAR@50\%} \\
\midrule
Falcon-H1-Tiny-90M          & 0.754 & 0.064 & 0.833 & 0.021 \\
Falcon3-7B-1.58bit          & 0.819 & 0.080 & 0.901 & 0.032 \\
Llama-3.2-3B                & 0.904 & 0.046 & 0.957 & 0.012 \\
Phi-mini-MoE-Instruct       & 0.904 & 0.054 & 0.964 & 0.017 \\
OLMo-2-1B                   & 0.827 & 0.058 & 0.890 & 0.024 \\
OLMo-2-1B-SFT               & 0.860 & 0.055 & 0.922 & 0.012 \\
OLMo-2-1B-Instruct          & 0.895 & 0.060 & 0.960 & 0.022 \\
Qwen3-4B-Instruct           & 0.963 & 0.024 & 0.993 & 0.005 \\
OLMo-3.1-32B-Instruct       & 0.947 & 0.023 & 0.988 & 0.002 \\
Qwen3.5-2B                  & 0.878 & 0.035 & 0.944 & 0.007 \\
Qwen3.5-9B                  & 0.943 & 0.032 & 0.965 & 0.005 \\
Qwen3.5-27B                 & 0.967 & 0.018 & 0.994 & 0.002 \\
\midrule
\textbf{Average}
& \textbf{0.888}
& \textbf{0.046}
& \textbf{0.943}
& \textbf{0.013} \\
\bottomrule
\end{tabular}%
}

\caption{Accuracy and FAR across confidence thresholds.}
\label{tab:selective_abstention}
\end{table}

\section{Long-Context Feature-Source Ablation}
\label{app:long_context}
Table~\ref{tab:long_context} shows that the final-token representation remains stronger than the document-level pooling alternatives as the number of retrieved documents increases, although performance generally decreases for longer contexts. The additional drop when a router trained at $k=5$ is evaluated on larger values of $k$ also indicates sensitivity to context-length distribution shift.

\begin{table}[th]
\centering
\small
\setlength{\tabcolsep}{3.5pt}
\caption{Long-context ablation as the number of retrieved documents increases.
Matched-$k$ trains and tests at the same context size, while train $k=5$
trains only on five-document contexts and evaluates on progressively longer
contexts. FAR and Conflict-F1 are reported at $k=20$.}
\label{tab:long_context}

\resizebox{\textwidth}{!}{%
\begin{tabular}{lllcccccc}
\toprule
\textbf{Model}
& \textbf{Setting}
& \textbf{Feature}
& \textbf{Acc. $k=5$}
& \textbf{Acc. $k=10$}
& \textbf{Acc. $k=15$}
& \textbf{Acc. $k=20$}
& \textbf{FAR $k=20$}
& \textbf{Conflict-F1 $k=20$} \\
\midrule

Qwen3-4B-Instruct
& matched-$k$
& Last token
& 0.858 & 0.842 & 0.804 & 0.821 & 0.113 & 0.843 \\

Qwen3-4B-Instruct
& matched-$k$
& Doc-end mean
& 0.575 & 0.554 & 0.454 & 0.500 & 0.312 & 0.639 \\

Qwen3-4B-Instruct
& matched-$k$
& Doc-token mean
& 0.637 & 0.550 & 0.533 & 0.454 & 0.325 & 0.608 \\

Llama-3.2-3B
& matched-$k$
& Last token
& 0.838 & 0.825 & 0.746 & 0.717 & 0.169 & 0.724 \\

Llama-3.2-3B
& matched-$k$
& Doc-end mean
& 0.588 & 0.537 & 0.487 & 0.517 & 0.312 & 0.616 \\

Llama-3.2-3B
& matched-$k$
& Doc-token mean
& 0.613 & 0.537 & 0.487 & 0.450 & 0.350 & 0.646 \\

\midrule

Qwen3-4B-Instruct
& train $k=5$
& Last token
& 0.858 & 0.846 & 0.808 & 0.808 & 0.181 & 0.792 \\

Llama-3.2-3B
& train $k=5$
& Last token
& 0.838 & 0.796 & 0.738 & 0.675 & 0.163 & 0.632 \\

\bottomrule
\end{tabular}%
}

\end{table}

\section{Random-Label Control}
\label{app:random_labels}

We also train the probe after randomly shuffling the class labels.
Table~\ref{tab:random_labels} shows that after randomly shuffling the class labels, average accuracy drops to $0.332 \pm 0.013$, which is close to chance for the three-way task. The large difference from the real-label results indicates that the probe performance is dependent on meaningful evidence-state labels rather than the classifier simply separating arbitrary classes. 

\begin{table}[t]
\centering
\small
\setlength{\tabcolsep}{6pt}
\caption{Random-label control. Real-label accuracy is taken from the main
results in Table~\ref{tab:main_results}; random-label accuracy is reported
as mean $\pm$ standard deviation.}
\label{tab:random_labels}

\begin{tabular}{lcc}
\toprule
\textbf{Model} & \textbf{Real-label Acc.} & \textbf{Random-label Acc.} \\
\midrule

Granite-3.1-8B-Instruct
& 0.88 & $0.333 \pm 0.012$ \\

OLMo-3.1-32B-Instruct
& 0.90 & $0.328 \pm 0.012$ \\

gpt-oss-20b
& 0.82 & $0.338 \pm 0.014$ \\

Qwen3.5-27B
& 0.91 & $0.330 \pm 0.009$ \\

Qwen3.5-2B
& 0.85 & $0.328 \pm 0.011$ \\

Qwen3.5-9B
& 0.91 & $0.326 \pm 0.019$ \\

Falcon-H1-Tiny-90M
& 0.69 & $0.328 \pm 0.024$ \\

Llama-3.2-3B
& 0.82 & $0.335 \pm 0.012$ \\

OLMo-2-1B
& 0.75 & $0.337 \pm 0.005$ \\

OLMo-2-1B-SFT
& 0.77 & $0.333 \pm 0.014$ \\

OLMo-2-1B-DPO
& 0.77 & $0.333 \pm 0.014$ \\

OLMo-2-1B-Instruct
& 0.78 & $0.332 \pm 0.011$ \\

Qwen3-4B-Instruct
& 0.89 & $0.341 \pm 0.010$ \\

\midrule
\textbf{Random-label average}
& -- & $\mathbf{0.332 \pm 0.013}$ \\

\bottomrule
\end{tabular}

\end{table}

\section{Binary vs.\ Three-Way Formulation}

\label{app:binary_ablation}

Table~\ref{tab:binary_threeway} presents the model-wise comparison between the binary and three-way formulations. The two settings have similar average accuracy, but the three-way formulation gives higher Macro-F1 and a lower FAR. This supports treating CONFLICT as a separate state, since doing so improves reliability without substantially reducing overall classification accuracy.

\begin{table}[th]
\centering
\small
\setlength{\tabcolsep}{2.8pt}
\caption{Comparison between the three-way and binary router formulations.
For FAR, lower values are better. The final column reports CONFLICT accuracy
for the three-way router.}
\label{tab:binary_threeway}

\resizebox{0.92\linewidth}{!}{%
\begin{tabular}{lccc|ccc|c}
\toprule
& \multicolumn{3}{c|}{\textbf{Three-Way}}
& \multicolumn{3}{c|}{\textbf{Binary}}
& \textbf{Three-Way} \\
\cmidrule(lr){2-4}
\cmidrule(lr){5-7}
\cmidrule(lr){8-8}

\textbf{Model}
& \textbf{Acc.}
& \textbf{Macro-F1}
& \textbf{FAR}
& \textbf{Acc.}
& \textbf{Macro-F1}
& \textbf{FAR}
& \shortstack{\textbf{CONFLICT}\\\textbf{Acc.}} \\

\midrule

Granite-3.1-8b-Ins
& 0.879 & 0.870 & 0.072
& 0.877 & 0.863 & 0.099
& 0.900 \\

OLMo-32B
& 0.904 & 0.904 & 0.062
& 0.900 & 0.887 & 0.072
& 0.919 \\

GPT-OSS
& 0.823 & 0.814 & 0.075
& 0.874 & 0.859 & 0.097
& 0.915 \\

Qwen-27B
& 0.912 & 0.902 & 0.080
& 0.896 & 0.884 & 0.091
& 0.919 \\

Qwen-2B
& 0.846 & 0.803 & 0.083
& 0.833 & 0.813 & 0.125
& 0.852 \\

Qwen-9B
& 0.907 & 0.887 & 0.071
& 0.881 & 0.866 & 0.096
& 0.892 \\

Falcon-tiny
& 0.696 & 0.684 & 0.138
& 0.757 & 0.713 & 0.140
& 0.733 \\

Llama-3.2-3b
& 0.816 & 0.814 & 0.093
& 0.805 & 0.783 & 0.156
& 0.856 \\

OLMo-2B-1b
& 0.752 & 0.731 & 0.143
& 0.747 & 0.717 & 0.198
& 0.756 \\

Qwen-3-4B
& 0.888 & 0.888 & 0.081
& 0.891 & 0.880 & 0.101
& 0.910 \\

\midrule

\textbf{Average}
& \textbf{0.829}
& \textbf{0.829}
& \textbf{0.104}
& \textbf{0.832}
& \textbf{0.811}
& \textbf{0.127}
& \textbf{0.850} \\

\bottomrule
\end{tabular}%
}

\end{table}

\section{Generation-Level Patching}
\label{app:generation_patching}
To test whether hidden-state interventions also affect the model's
autoregressive output, we additionally perform generation-level patching.
The model receives the same RAG prompt used in the main experiments and
generates a free-form response after intervention at the selected layer.

Across models and patching directions, 63--83\% of patched generations have
Jaccard divergence $\geq 0.30$ from their corresponding unpatched
generations, while the degeneration rate is $\leq 0.01$. Sentence-level
comparisons further show a semantic-similarity gain of $+0.47$ to $+0.93$
over the within-class control, with BERTScore F1 $\geq 0.87$.

\section{Enlarged Figures}
\label{app:zoom}

\begin{figure}[htbp]
    \centering
    \includegraphics[width=0.9\linewidth]{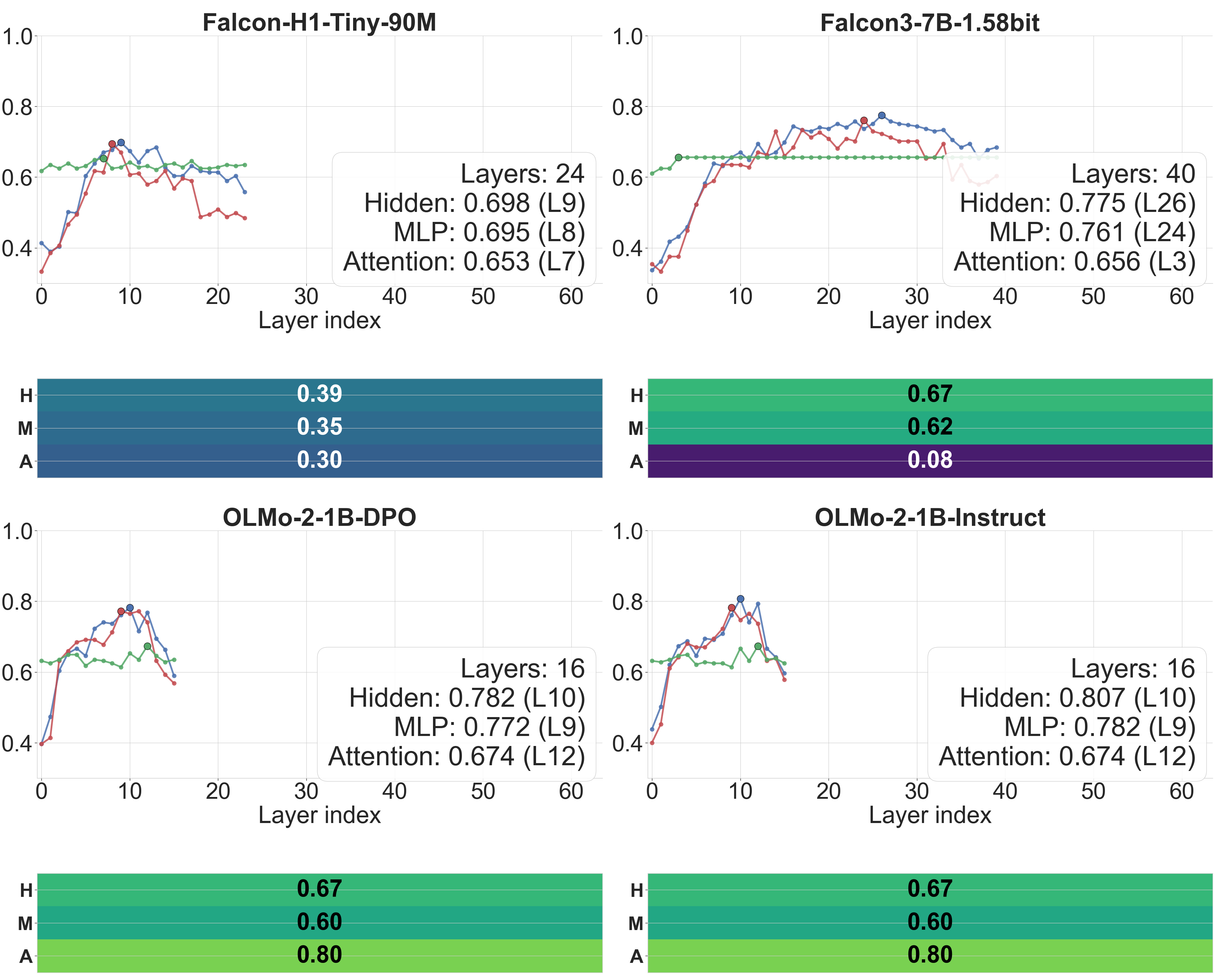}

    \vspace{1em}

    \includegraphics[width=0.9\linewidth]{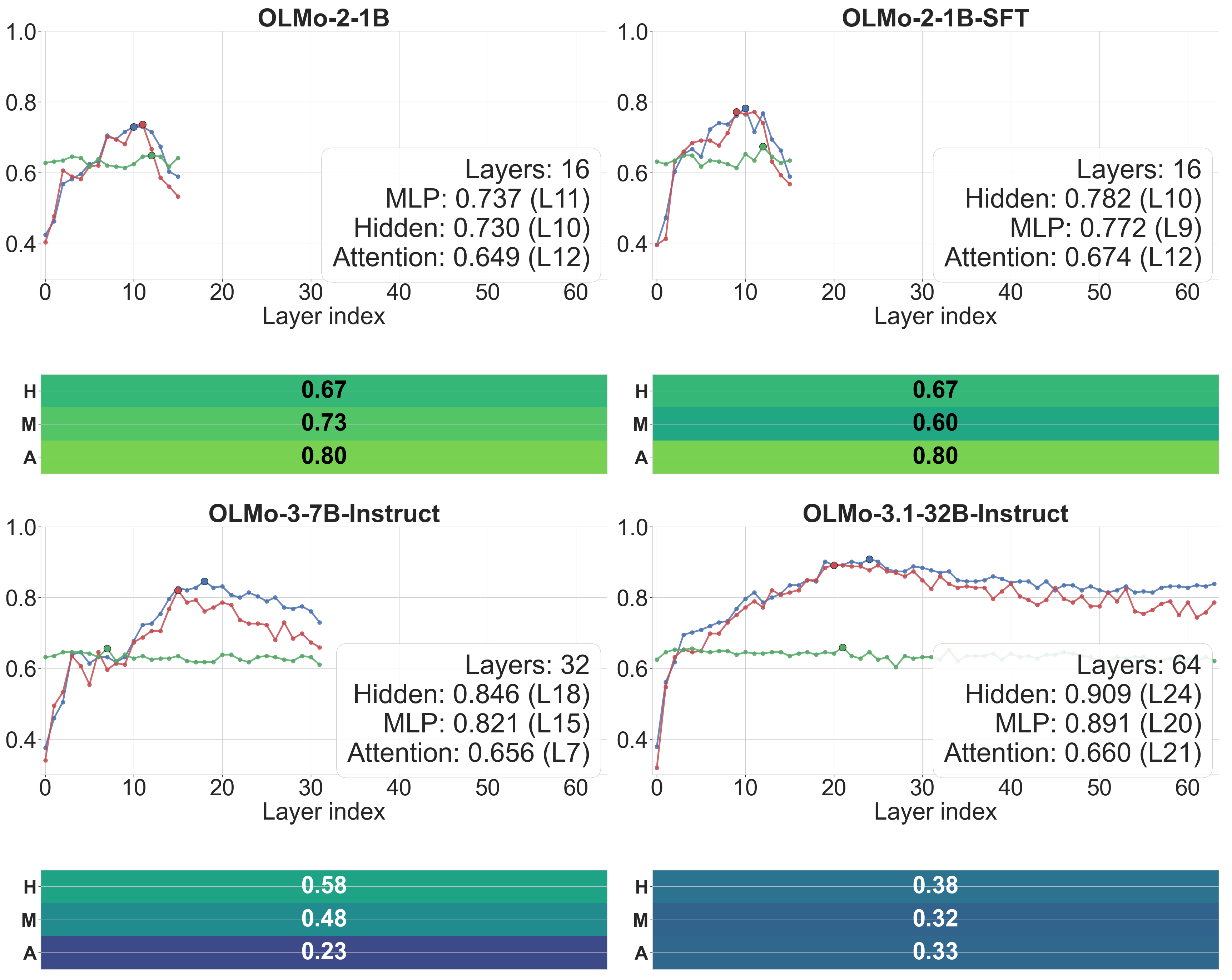}
    \caption{Zoomed layer-wise validation accuracy}
    \label{fig:zoomed_acc}
\end{figure}

\begin{figure}[htbp]\ContinuedFloat
    \centering
    \includegraphics[width=0.9\linewidth]{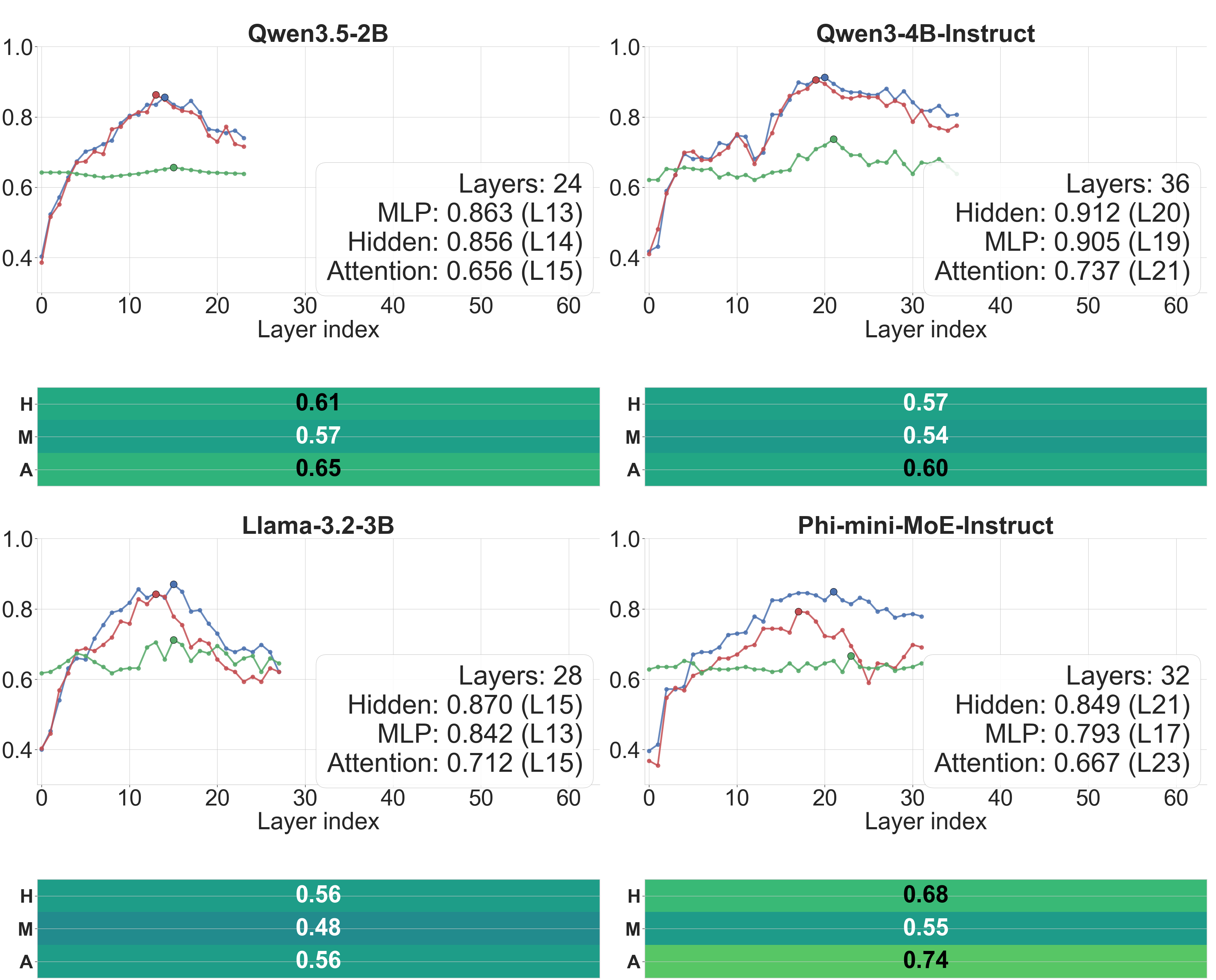}

    \vspace{1em}

    \includegraphics[width=0.9\linewidth]{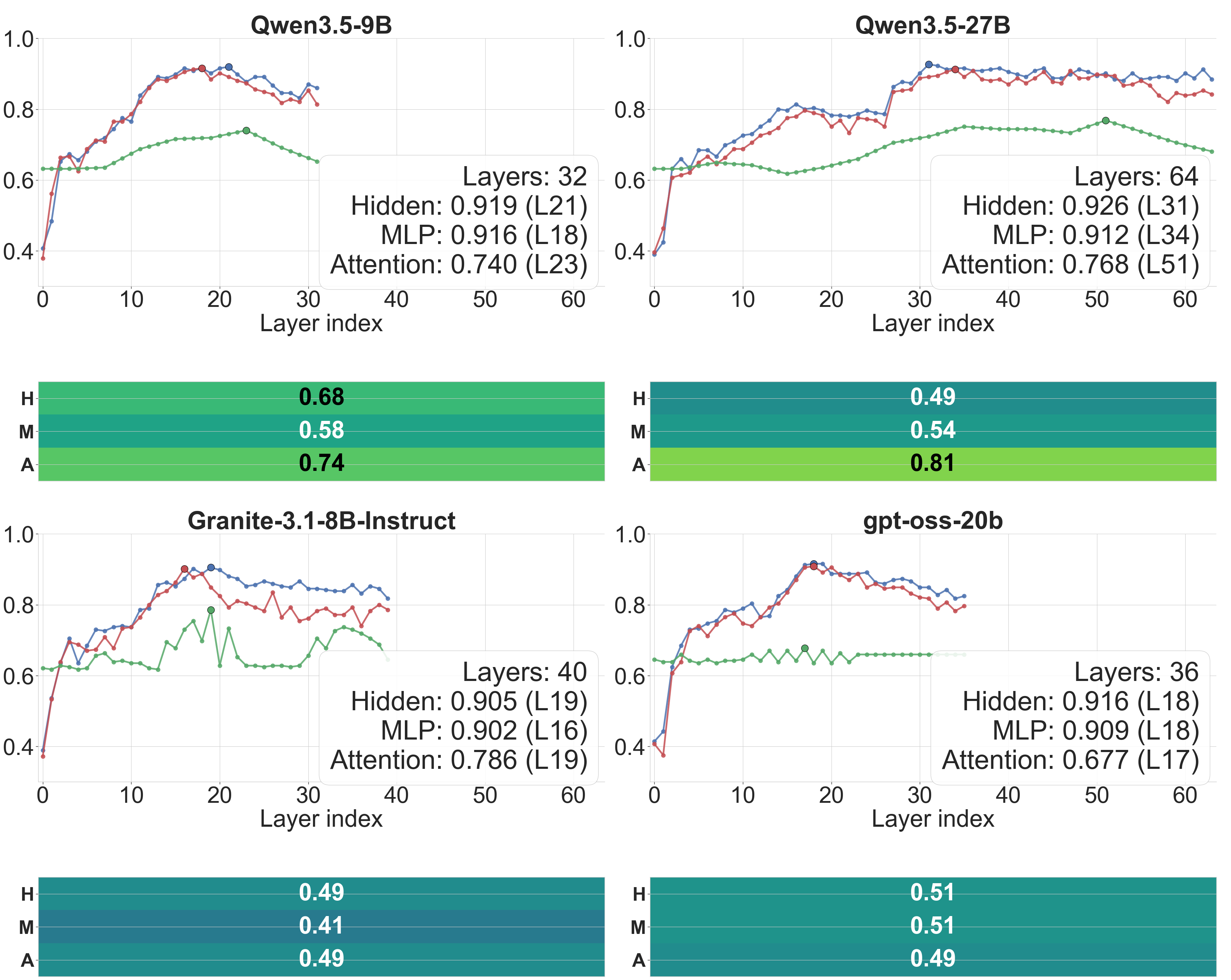}
    \caption{Zoomed layer-wise validation accuracy (continued)}
\end{figure}

\end{document}